\documentclass[letterpaper]{article} 
\usepackage{aaai25}  
\usepackage{times}  
\usepackage{helvet}  
\usepackage{courier}  
\usepackage[hyphens]{url}  
\usepackage{graphicx} 
\usepackage{natbib}  
\usepackage{caption} 
\usepackage{algorithm}
\usepackage{algorithmic}
\usepackage{amssymb}
\usepackage{amsmath}
\usepackage{multirow}
\usepackage{booktabs}
\usepackage{newfloat}
\usepackage{listings}
\DeclareCaptionStyle{ruled}{labelfont=normalfont,labelsep=colon,strut=off} 
\floatstyle{ruled}
\newfloat{listing}{tb}{lst}{}
\floatname{listing}{Listing}
\title{MixNet: Efficient Global Modeling for Ultra-High-Definition Image Restoration}
\author{
    Chen Wu\textsuperscript{\rm 1}, Zhuoran Zheng\equalcontrib, Yuning Cui, Wenqi Ren
}
\affiliations{
    \textsuperscript{\rm 1}University of Science and Technology of China\\
    \textsuperscript{\rm 2}Sun Yat-sen University\\
    \textsuperscript{\rm 3}Technical University of Munich\\
    wuchen5X@mail.ustc.edu.cn
    
}

\usepackage{bibentry}

\begin{document}

\maketitle

\begin{abstract}
Recent advancements in image restoration methods employing global modeling have shown promising results. However, these approaches often incur substantial memory requirements, particularly when processing ultra-high-definition (UHD) images. 
In this paper, we propose a novel image restoration method called MixNet, which introduces an alternative approach to global modeling approaches and is more effective for UHD image restoration. 
To capture the long-range dependency of features without introducing excessive computational complexity, we present the Global Feature Modulation Layer (GFML). GFML associates features from different views by permuting the feature maps, enabling efficient modeling of long-range dependency. In addition, we also design the Local Feature Modulation Layer (LFML) and Feed-forward Layer (FFL) to capture local features and transform features into a compact representation. This way, our MixNet achieves effective restoration with low inference time overhead and computational complexity. 
We conduct extensive experiments on four UHD image restoration tasks, including low-light image enhancement, underwater image enhancement, image deblurring and image demoiréing, and the comprehensive results demonstrate that our proposed method surpasses the performance of current state-of-the-art methods. The code will be available at \url{https://github.com/5chen/MixNet}.
\end{abstract}

%

\section{Introduction}
In recent years, high-resolution imaging has experienced significant advancements due to the emergence of sophisticated imaging sensors and displays. This has led to a rapid evolution of Ultra-High-Definition (UHD) imaging technology. However, the inherent high-resolution characteristics of UHD images make them more susceptible to noise during the imaging process. Additionally, the increased number of pixels in UHD images challenges the efficiency of existing image processing methods. Therefore, recovering clear, realistic, and clean images from degraded UHD images (\textit{e.g.}, low-light, blur) is an extremely challenging task.

Although many learning-based restoration methods have made significant progress, they always focus on low-resolution images~\cite{restormer,wang2022uformer,yang2023dual}. For example, in the LLIE task, the most common dataset used is LOL~\cite{LOL}, which consists of images with a resolution of $600 \times 400$. When faced with UHD images, these SOTA methods often encounter memory overflow and cannot perform full-resolution inference on consumer-grade GPUs. This limitation may potentially impact the applications of UHD imaging devices.

\begin{figure}[t!]
    \centering
    \includegraphics[width=\linewidth]{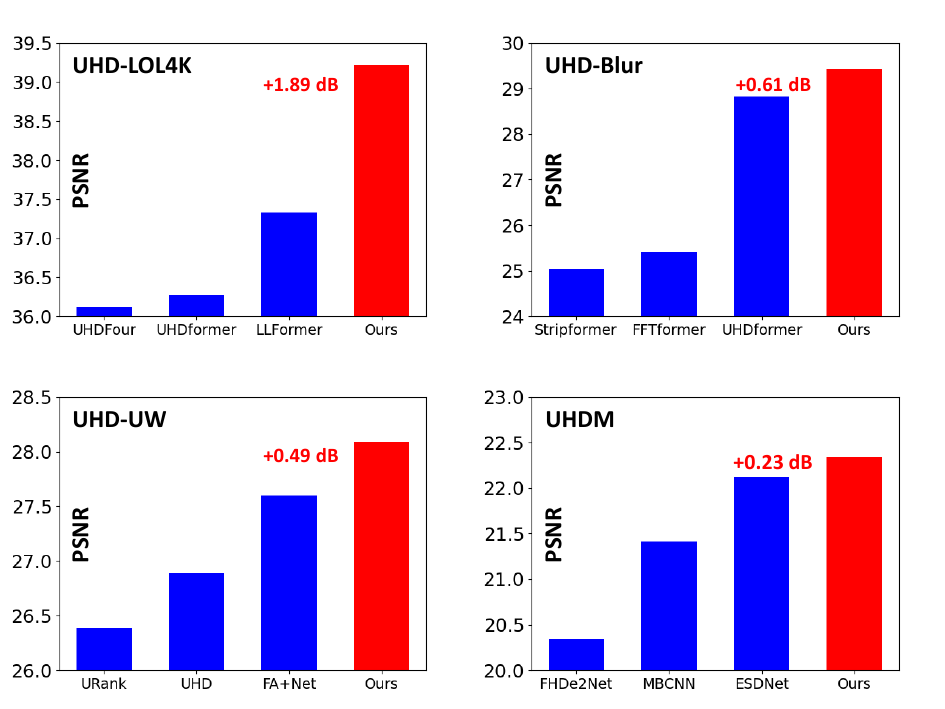}
    \caption{Our MixNet significantly outperforms state-of-the-art UHD image restoration methods, including UHD, UHDFour, UHDformer and ESDNet, on four UHD image restoration benchmarks.}
    \label{fig:bar}
\end{figure}

Recently, there have been some explorations into restoration methods for UHD images~\cite{LLFormer,UHDFour,uhdformer,ultradehaze,uhdhdr}. UHD~\cite{ultradehaze} proposes a multi-guided bilateral upsampling model for UHD image dehazing. The core idea of it is to learn the local affine coefficients of the bilateral grid from the low-resolution image and apply them to the full-resolution image. UHDFour~\cite{UHDFour} attempts to perform UHD low-light image enhancement in the Fourier domain and achieves good results on the real-world dataset. These methods are mainly based on CNN and overlook the modeling of global features. LLFormer~\cite{LLFormer} and UHDformer~\cite{uhdformer} both attempt to use transformers to capture the long-range dependency of features. The excellent long-range dependency capture capability of the transformer undoubtedly brings excellent results. However, LLFormer is also unable to perform full-resolution inference of UHD images on consumer-grade GPUs. It requires the image to be divided into multiple patches without overlapping for inference, and then the results are stitched together. This process undoubtedly degrades the quality of image restoration and may introduce boundary artifacts. While UHDFormer chooses to model the features globally in the 8$\times$ downsampled low-resolution space, this undoubtedly leads to the loss of too much high-frequency information, which is crucial for restoring the high-quality image. Furthermore, the inference time overhead introduced by the transformer architecture has been a persistent concern. 

In this paper, we propose a simple but effective UHD image restoration method, called MixNet. Specifically, we design the Feature Mixing Block which mainly contains Global Feature Modulation Layer (GFML), Local Feature Modulation Layer (LFML), and Feed-forward Layer (FFL). GFML is the key component of MixNet, enabling efficient modeling of long-range dependency. The use of self-attention in transformers endows them with powerful long-range dependency modeling capabilities~\cite{attention}. However, this approach comes with significant computational overhead. Inspired by the vanilla MLP-Mixer~\cite{MlpMixer}, but different from it, GFML encodes the feature maps of images from the perspectives of the width $W$, height $H$, and channel $C$. Through simple dimensional transformation operations, the information encoded in each feature map is associated and fused. It is worth noting that dimension transformation operations do not introduce additional parameters, which means that by permuting the feature maps from different views, GFML achieves efficient modeling of long-range dependency. LFML is designed to capture the local features of the images. LFML adaptively calculates a set of parameters for the redistribution of weights in the channel dimension, emphasizing the importance of specific channels. The results of GFML and LFML are fed to FFL which aim to transform features to a compact representation. This way, our MixNet achieves a better trade-off between model complexity and performance. Extensive experiments on both synthetic and real-world datasets demonstrate that MixNet surpasses existing SOTA methods in terms of restoration quality and generalization ability (See Fig.~\ref{fig:bar}).

Our contributions can be summarized below:
\begin{itemize}
    \item We propose a simple but effective UHD image restoration method called MixNet. MixNet can perform full-resolution inference of UHD images on consumer-grade GPUs and achieve high-quality UHD image restoration. 
    \item We develop a novel efficient feature modulation mechanism to capture long-range dependency, which does not need to project the input image from 3D to 2D; it maintains the spatial properties between pixels by modeling directly on 3D space.
    \item We evaluate the proposed method quantitatively and qualitatively on synthetic and real-world datasets, and the results show that our MixNet achieves a favorable trade-off between accuracy and model complexity.
\end{itemize}

\section{Related Work}
\subsection{UHD Image Restoration}
UHD image restoration has been an emerging topic in recent years~\cite{liang2021high,zamfir2023towards,guan2022memory}. In ~\cite{ultradehaze,uhdhdr}, the authors introduced bilateral learning to achieve dehazing of the UHD image and HDR reconstruction of the UHD image. The core idea is to learn the local affine coefficients of the bilateral grid from the low-resolution image and apply them to the full-resolution image. UHD-SFNet~\cite{uhdUnderwater} and FourUHD~\cite{UHDFour} explore underwater UHD image enhancement and UHD low-light image enhancement in the Fourier domain. They are all inspired by the discovery of most luminance information concentrates on amplitudes. Due to the unavoidable need for downsampling in existing UHD image restoration methods, NSEN~\cite{NSEN} proposes a spatial-variant and invertible nonuniform downsampler that adaptively adjusts the sampling rate according to the richness of details. LLFormer~\cite{LLFormer} is the first transformer-based method to complete UHD-LLIE task, but like most previous methods, it is unable to perform full-resolution inference on consumer-grade GPUs. To avoid this issue, UHDformer~\cite{uhdformer} designs two learning spaces, by building feature transformation from a high-resolution space to a low-resolution one, to compute self-attention in the low-resolution space, which significantly reduces the computational overhead. Despite the promising results achieved by the aforementioned approaches, there is still room for a favorable trade-off between reconstruction performance and model efficiency, prompting us to develop a more efficient global modeling method.

\begin{figure*}[t!]
    \centering
    \includegraphics[width=\linewidth]{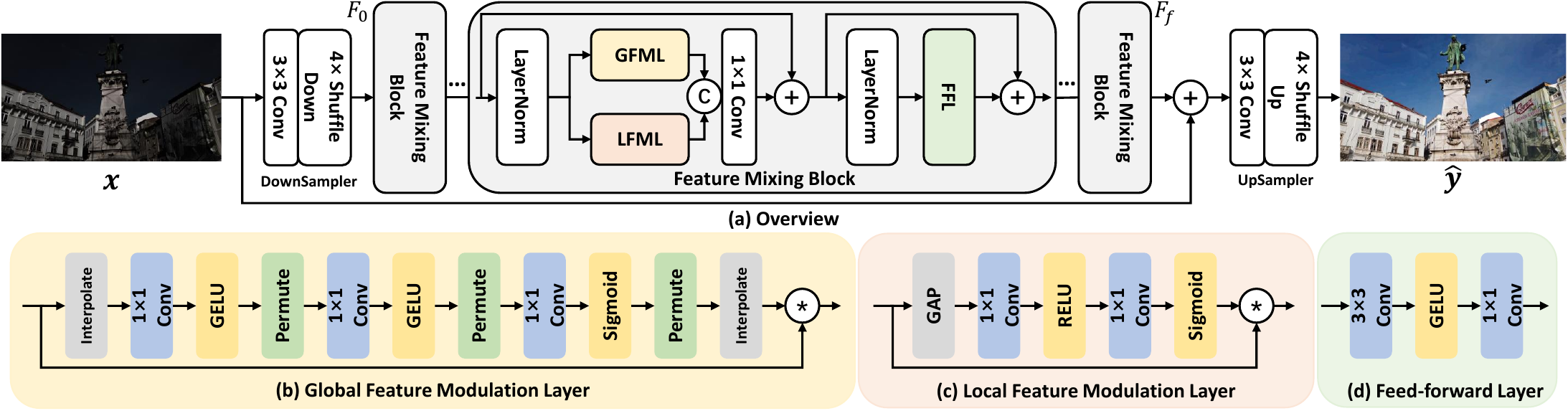}
    \caption{The overall architecture of the proposed MixiNet. MixNet first transforms the input degraded image into the feature space using a DownSampler, performs feature extraction using a series of Feature Mixing Blocks (FMBs), and then reconstructs these extracted features using an UpSampler. The FMB mainly contains a Global Feature Modulation Layer (GFML), a Local Feature Modulation Layer (LFML), and a Feed-forward Layer (FFL).}
    \label{fig:framework}
\end{figure*}

\subsection{Global Modeling}
In recent years, global modeling techniques have been receiving increasing attention in the field of computer vision, especially after the great success of Vision Transformer~\cite{vit} in the visual domain. Transformer learns long-range dependencies between image patch sequences for global-aware modeling, and the interaction of these non-local features can significantly improve network performance. Various image restoration
algorithms based on transformer have been proposed and achieve superior performance in restoration tasks such as image deraining~\cite{chen2023learning,li2023dilated}, image dehazing~\cite{song2023vision,qiu2023mb}, low-light image enhancement~\cite{zhao2021deep,Uretinexnet,SNRNet,Retinexformer} and image deblurring~\cite{Stripformer,fftformer}. However, transformer-based methods are still difficult to apply directly to UHD image restoration tasks due to their huge computation cost. Owing to the low time complexity and hardware-friendly nature of MLP, some recent studies~\cite{MlpMixer} have explored the use of MLP for global modeling. However, MLP-based methods inevitably lose the spatial information between pixels during global modeling, which is unfavorable for pixel-level image restoration tasks. In contrast, our approach encodes the feature maps of images from the perspectives of width $W$, height $H$, and channel $C$, preserving spatial information while achieving efficient global modeling.

\section{Methodology}
MixNet aims to recover clean and realistic UHD images from degraded images which often exhibit significantly reduced visibility, low contrast, and high levels of noise. We provide the overall pipeline of our method and further details on the critical components of our approach below.

\subsection{Overview}
An overview of MixNet is shown in Fig.~\ref{fig:framework} (a). For a given degraded image ${x}\in\mathbb{R}^{H\times W\times3}$, we first map it to the feature space through a DownSampler to obtain the low-level features ${F}_{0}\in\mathbb{R}^{H \times W \times C}$, where $H$, $W$, and $C$ represent height, width, and channel, respectively. Then, the multiple stacked Feature Mixing Blocks (FMBs) are used to generate finer deep features ${F}_{i}$ from ${F}_{0}$ for normal-light image reconstruction, where a FMB has a Global Feature Modulation Layer (GFML), a Local Feature Modulation Layer (LFML) and a Feed-forward Layer (FFL). Lastly, the sum of the final features ${F}_{f}$ and ${F}_{0}$ is fed to an UpSampler to obtain the restored image ${\hat{y}}$.

\begin{figure}[t!]
    \centering
    \includegraphics[width=\linewidth]{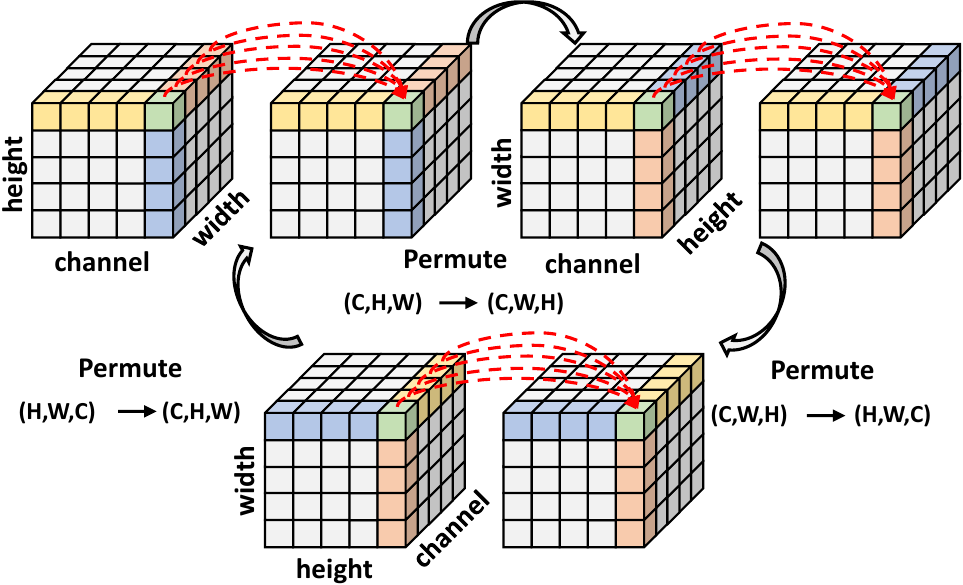}
    \caption{Visualization of dimensional transformation operations. MixNet can capture long-range dependency of features with few parameters by employing simple dimension transformation operations.}
    \label{fig:mixer}
\end{figure}

\subsection{Global Feature Modulation Layer}
Recent studies suggest that the notable performance of transformers across diverse tasks stems from their implementation of the key multi-head self-attention (MHSA) mechanism~\cite{attention}. MHSA enables the model with the capability of long-range feature interaction and dynamic spatial weighting, both of which contribute to get promising results.
However, this mechanism comes with significant computational overhead. To capture the long-range dependency of features, we ingeniously establish long-range relationships in both spatial and channel dimensions by employing simple dimension transformation operations. This allows for achieving long-range dependency of features by only using a few parameters.

The architecture of the GFML is shown in Fig.~\ref{fig:framework} (b) and the more details of dimension transformation operations are shown in Fig.~\ref{fig:mixer}. Specifically, we first adjust the resolution of the normalized input features, and then perform some dimension transformation operations on them. Given the input feature $F_{in}$, this procedure can be formulated as:
\begin{align}
    F_{t}&=\operatorname{IP}(F_{in}),\\
    \hat{F}_{t}&=\underbrace{\operatorname{Permute}(\operatorname{GELU}(\operatorname{Conv_{1\times1}}(F_{t})))}_{\times3},
\end{align}
where $F_{t}$ and $\hat{F}_{t}$ are intermediate results. $\operatorname{IP}(\cdot)$ corresponds to the interpolation operation, $\operatorname{Conv_{1\times1}}(\cdot)$ is a $1\times1$ convolution, $\operatorname{GELU}(\cdot)$ represents GELU function, $\operatorname{Permute}(\cdot)$ denotes the dimension transformation operation and $\times 3$ denotes three operations in sequence. It is worth noting that before executing the last dimension transformation operation, we replace the GELU function with the Sigmoid function.

\begin{figure*}[t!]
    \centering
    \includegraphics[width=\linewidth]{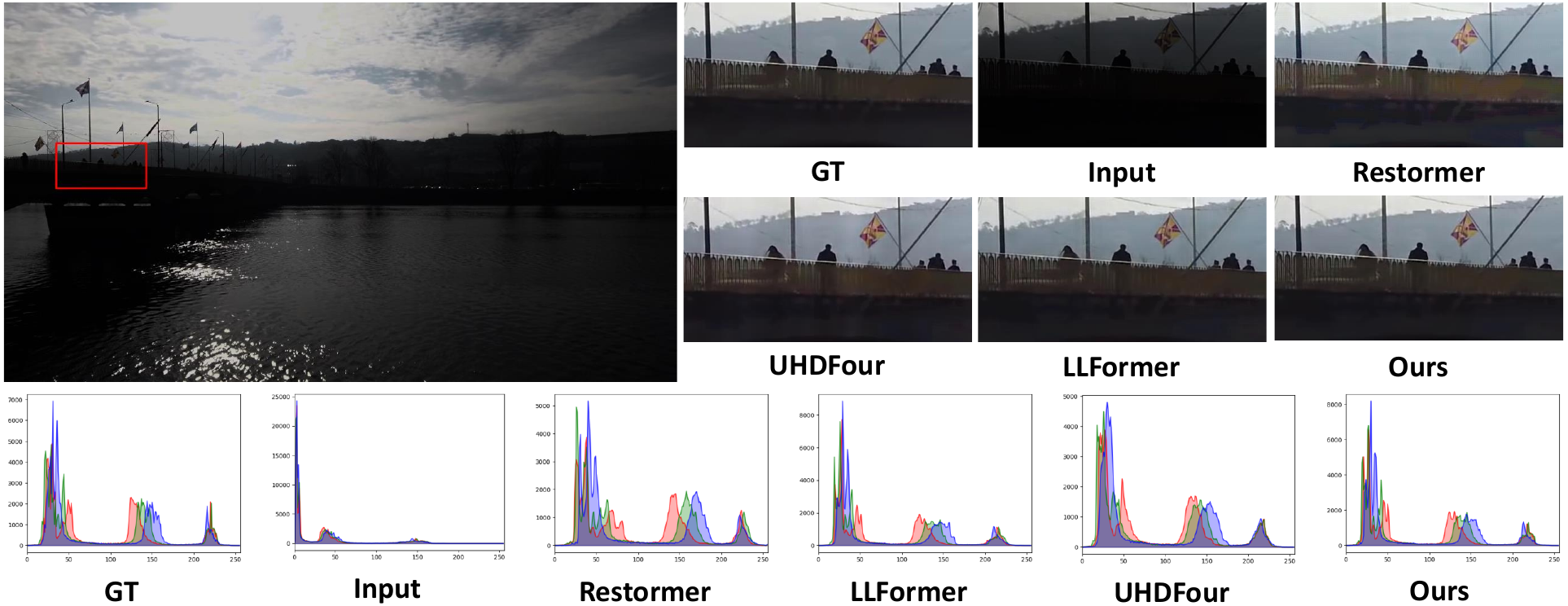}
    \caption{Visual quality comparisons with state-of-the-art methods on UHD-LOL4K dataset. The last row shows the color histogram of the image. Our method has the closest color to GT image.}
    \label{fig:lle}
\end{figure*}

Afterward, We use interpolation to adjust the feature $\hat{F}_{t}$ to its original resolution to estimate the attention map and adaptively modulate the input $F_{in}$ according to the estimated attention via element-wise product. This process can be written as:
\begin{equation}
    F_{g}=\operatorname{IP}(\hat{F}_{t})\odot F_{in},
\end{equation}
where $F_{g}$ are the final output features and $\odot$ represents element-wise product.

\subsection{Local Feature Modulation Layer}
LFML aims to capture the local feature of images and the architecture of it is shown in Fig.~\ref{fig:framework} (c). For a set of feature maps, their impact on the final results is different~\cite{CA}, and we hope the model can focus on the important feature maps. We first feed the normalized input features into global average pooling to shift the focus of the model from space dimension to channel dimension, and then feed them into a series of convolutions to get deep features. Finally, these deep features are processed by the sigmoid function to obtain the weight of each channel. The normalized input features adaptively adjust the importance of the channel based on these weights via element-wise product. The LFML can be written as:
\begin{align}
    F_{w}&=\operatorname{Sigmoid}(\operatorname{Conv}(\operatorname{ReLU}(\operatorname{Conv}(\operatorname{GAP}(F_{in}))))),\\
    F_{l}&=F_{w}\odot F_{in},
\end{align}
where $F_{in}$ stand for the input features, $F_{w}$ are the channel weights and $F_{l}$ denote the final output features. $\operatorname{Sigmoid}(\cdot)$ represents Sigmoid function, $\operatorname{Conv}(\cdot)$ is a $1\times1$ convolution and $\operatorname{GAP}(\cdot)$ denotes global average pooling.

\subsection{Feed-forward Layer}
Through GFML and LFML, we obtain global features and local features respectively. Undoubtedly, there will be overlapping parts in the feature space between these two, leading to redundant information. To transform features into a compact representation, we introduce FFL into our model. As shown in Fig.~\ref{fig:framework} (d), FFL comprises a $3\times3$ convolution, a $1\times1$ convolution, and a GELU function. Within this, the first $3\times3$ convolution encodes the spatially local contexts and doubles the number of channels of the input features for mixing channels; the later $1\times1$ convolution reduces the channels back to the original input dimension. In this way, the model learns a more compact representation of features and ignores some unimportant information~\cite{tan2021efficientnetv2}. The FFL can be expressed as:
\begin{equation}
    F_{out}=\operatorname{Conv_{1\times 1}(\operatorname{GELU}(\operatorname{Conv}_{3\times 3}((F_{in})))))},
\end{equation}
where $F_{in}$ and $F_{out}$ are the input features and output features, respectively.

\subsection{Feature Mixing Block}
Here, we describe our FMB pipeline in general, which can be written as:
\begin{align}
\hat{F}_{in}&=\operatorname{Conv}[\operatorname{GFML}(\operatorname{LN}(F_{in}));\operatorname{LFML}(\operatorname{LN}(F_{in}))]+F_{in},\\
F_{out}&=\operatorname{FFL}(\operatorname{LN}(\hat{F}_{in}))+\hat{F}_{in},
\end{align}
where $F_{in}$ stand for the input features, $\hat{F}_{in}$ are the intermediate features and $F_{out}$ denote the output features. [;] is concatenation operation, and $\operatorname{Conv}$ respresents the $1\times 1$ convolution.

\subsection{Loss Functions}
To optimize the weights and biases of the network, we utilize the L1 loss in the RGB color space as the basic reconstruction loss:
\begin{equation}
    L=\|y-\hat{y}\|_1,
\end{equation}
where $y$ and $\hat{y}$ denote the ground truth and restored image, respectively.

\begin{figure*}[t!]
    \centering
    \includegraphics[width=\linewidth]{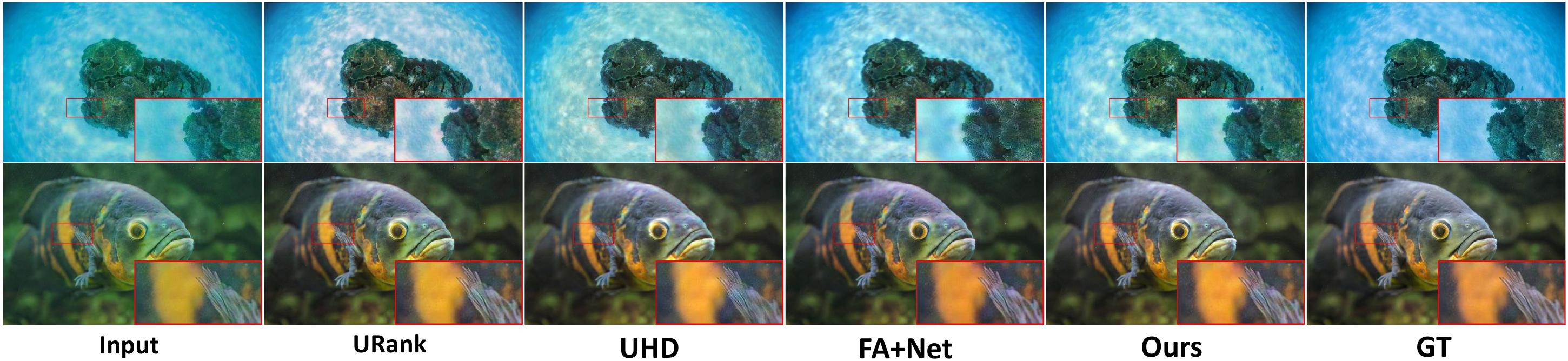}
    \caption{Visual quality comparisons with state-of-the-art methods on UHD-UW dataset. Please zoom in for details.}
    \label{fig:uw}
\end{figure*}
\begin{table}[t!]
\resizebox{\linewidth}{!}{
\begin{tabular}{c|c|c|cc|c}
\toprule[0.5mm]
Methods   & Type                                          & Venue    & PSNR $\uparrow$  & SSIM $\uparrow$ & Param $\downarrow$ \\ \hline
Z\_DCE++  & \multicolumn{1}{c|}{\multirow{4}{*}{non-UHD}} & TPAMI'21 & 15.58 & 0.934 & 10.56K\\
RUAS      & \multicolumn{1}{c|}{}                         & CVPR'21  & 14.68 & 0.757 & 3.44K\\
Uformer   & \multicolumn{1}{c|}{}                         & CVPR'22  & 29.98 & 0.980 & 20.63M\\
Restormer & \multicolumn{1}{c|}{}                         & CVPR'22  & 36.90 & 0.988 & 26.11M\\ \hline
NSEN      & \multicolumn{1}{c|}{\multirow{5}{*}{UHD}}     & MM'23    & 29.49 & 0.980 & 2.67M\\
UHDFour   & \multicolumn{1}{c|}{}                         & ICLR'23  & 36.12 & \underline{0.990} & 17.54M\\
LLFormer  & \multicolumn{1}{c|}{}                         & AAAI'23  & \underline{37.33} & 0.988 & 24.52M \\
UHDformer & \multicolumn{1}{c|}{}                         & AAAI'24  & 36.28 & 0.989 & 339.3K \\
MixNet    & \multicolumn{1}{c|}{}                         & -        & \textbf{39.22} & \textbf{0.992} & 7.77M\\
\bottomrule[0.5mm]
\end{tabular}}
\caption{Comparison of quantitative results on UHD-LOL4K dataset. Best and second best values are indicated with \textbf{bold} text and \underline{underlined} text respectively.}
\label{tab:UHDLOL}
\end{table}

\section{Experiments}
\subsection{Implementation Details and Datasets}
\subsubsection{Implementation Details.}
We conduct experiments in PyTorch on six NVIDIA GeForce RTX 3090 GPUs. To optimize the network, we employ the Adam optimizer with a initial learning rate $2 \times 10^{-4}$ and a cosine annealing strategy is used for the decay of the learning rate. We randomly crop the full-resolution 4K image to a resolution of $512\times512$ as the input and the batch size is set ot 24. Except for the UHD underwater image enhancement task, where the dataset is relatively small, we only perform 100K iterations. For all other tasks, we perform 300K iterations. To augment the training data, random horizontal and vertical flips are applied to the input images. The number of FMB and feature channels is set to 8 and 48, respectively.

\subsubsection{Datasets.}
To verify the effectiveness of MixNet, we evaluate it on the UHD-LOL dataset~\cite{LLFormer}, UHD-Blur dataset~\cite{uhdformer}, UHD-UW dataset~\cite{uhdUnderwater} and UHDM dataset~\cite{UHDM}. UHD-LOL consists of two subsets, namely UHD-LOL4K and UHD-LOL8K, which contain UHD images with resolutions of 4K and 8K in low-light conditions, respectively. In this study, we focus on the UHD-LOL4K subset to validate the effectiveness of WowNet. The UHD-LOL4K subset comprises a total of 8,099 image pairs, with 5,999 pairs allocated for training purposes and 2,100 pairs designated for testing. UHD-Blur dataset is re-collected from the datasets of Deng \textit{et al.}~\cite{deng2021multi} by Wang \textit{et al.}~\cite{uhdformer} and it has 1,964 blur-clear image pairs for training and 300 pairs for testing. UHD-UW dataset is our organization of the dataset proposed in UHD-SFNet~\cite{uhdUnderwater}, where we remove the images with resolutions lower than $512\times512$, and sequentially select 80 images as our training set, with the remaining 13 images as our test set. UHDM dataset is a real-world 4K resolution demoir´eing dataset, consisting of 5,000 image pairs in total, with 4500 images used for training and 500 images used for testing. 

\begin{table}[t!]
\resizebox{\linewidth}{!}{
\begin{tabular}{c|c|c|cc|c}
\toprule[0.5mm]
Methods   & Type                                          & Venue    & PSNR $\uparrow$ & SSIM $\uparrow$ & Param $\downarrow$ \\ \hline
Shallow-uwnet     & \multicolumn{1}{c|}{\multirow{5}{*}{non-UHD}}    & AAAI'21   & 23.36 & 0.798 & 0.22M\\
USUIR      & \multicolumn{1}{c|}{}                         & AAAI'22  & 24.13 & 0.834 &  3.36M\\
PUIE-Net  & \multicolumn{1}{c|}{} & ECCV'22 & 25.77 & 0.859 & 1.40M\\
URank & \multicolumn{1}{c|}{}                         & AAAI'23  & 26.39 &  0.911 & 3.15M\\
FA+Net   & \multicolumn{1}{c|}{}                         & BMVC'23  & \underline{27.60} & \underline{0.916} & 9K\\ \hline
UHD      & \multicolumn{1}{c|}{\multirow{3}{*}{UHD}}     & ICCV'21    & 26.89 & 0.907 & 34.5M\\
UHD-SFNet  & \multicolumn{1}{c|}{}     & ACCV'22    & 26.31 & 0.914 & 37.31M\\
Ours      & \multicolumn{1}{c|}{}                         & -        & \textbf{28.09} & \textbf{0.932} & 7.77M\\
\bottomrule[0.5mm]
\end{tabular}}
\caption{Comparison of quantitative results on UHD-UW dataset. Best and second best values are indicated with \textbf{bold} text and \underline{underlined} text respectively.}
\label{tab:UHDUW}
\end{table}

\subsection{Comparisons with the State-of-the-art Methods}
\subsubsection{Evaluation.}
Two well-known metrics, Peak Signal-to-Noise Ratio (PSNR) and Structural Similarity (SSIM)~\cite{ssim}, are employed for quantitative comparisons. Higher values of these metrics indicate superior performance of the methods. PSNR and SSIM are calculated along the RGB channels. We also report the trainable parameters (Param) and running time (RT) which you can find in Tab.~\ref{tab:time}. For methods that cannot perform full-resolution inference, we adopt the previous patch-based or downsample-based approaches to handle them.

\begin{figure*}[t!]
    \centering
    \includegraphics[width=\linewidth]{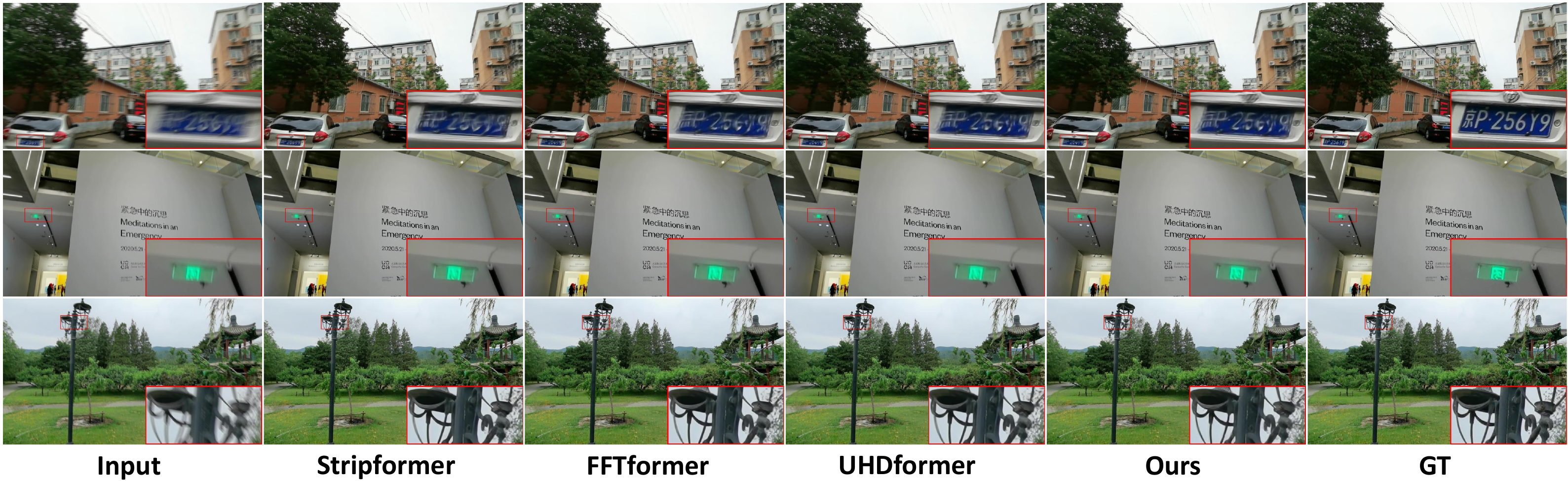}
    \caption{Visual quality comparisons with state-of-the-art methods on UHD-Blur dataset. Please zoom in for details.}
    \label{fig:blur}
\end{figure*}
\begin{table}[t!]
\resizebox{\linewidth}{!}{
\begin{tabular}{c|c|c|cc|c}
\toprule[0.5mm]
Methods   & Type                                          & Venue    & PSNR $\uparrow$  & SSIM $\uparrow$ & Param $\downarrow$ \\ \hline
MIMO-Unet++  & \multicolumn{1}{c|}{\multirow{5}{*}{non-UHD}} & ICCV'21 & 25.03 & 0.752 & 16.1M\\
Restormer      & \multicolumn{1}{c|}{}                         & CVPR'22  & 25.21 & 0.752 & 26.1M\\
Uformer     & \multicolumn{1}{c|}{}                         & CVPR'22   & 25.27 & 0.752 & 20.6M\\
Stripformer   & \multicolumn{1}{c|}{}                         & ECCV'22  & 25.05 & 0.750 & 19.7M\\
FFTformer & \multicolumn{1}{c|}{}                         & CVPR'23  & 25.41 & 0.757 & 16.6M\\ \hline
UHDformer      & \multicolumn{1}{c|}{\multirow{2}{*}{UHD}}     & AAAI'24    & \underline{28.82} & \underline{0.844} & 0.34M\\
Ours      & \multicolumn{1}{c|}{}                         & -        & \textbf{29.43} & \textbf{0.855} & 5.22M\\
\bottomrule[0.5mm]
\end{tabular}}
\caption{Comparison of quantitative results on UHD-Blur dataset. Best and second best values are indicated with \textbf{bold} text and \underline{underlined} text respectively.}
\label{tab:UHDBlur}
\end{table}

\subsubsection{Low-light UHD Image Enhancement Results.}
For low-light UHD image enhancement task, we compare our MixNet with approaches Z\_DCE++~\cite{zero++}, RUAS~\cite{RUAS}, ELGAN~\cite{ELGAN}, Uformer~\cite{wang2022uformer}, Restormer~\cite{restormer}, NSEN~\cite{NSEN}, UHDFour~\cite{UHDFour}, LLFormer~\cite{LLFormer} and UHDformer~\cite{uhdformer}. As shown in Tab.~\ref{tab:UHDLOL}, our method outperforms the current state-of-the-art method, LLFormer, by 1.89dB in PSNR, which is undoubtedly a significant improvement. And you can find the visualization results in Fig.~\ref{fig:lle}. From the color histograms, it can be seen that the images restored using our method have the color distribution closest to the ground truth.

\subsubsection{Underwater UHD Image Enhancement Results.}
We evaluate underwater UHD image enhancement results on UHD-UW dataset and compare our method with several methods~\cite{Shallow-uwnet,USUIR,PUIENet,URank,FA+Net,ultradehaze,uhdUnderwater}. The results are shown in Tab.~\ref{fig:uw}, and compared with recent state-of-the-art FA+Net~\cite{FA+Net}, MixNet achieves a 0.49dB performance improvement in PSNR. You can see the comparison of the visualization results in Fig.~\ref{fig:uw}, which shows that our method achieves better color correction.

\subsubsection{UHD Image Deblurring Results.} 
In the UHD image deblurring task, we evaluate our proposed MixNet against existing deblurring approaches, including MIMO-Unet++~\cite{MIMO}, Restormer~\cite{restormer}, Uformer~\cite{wang2022uformer}, Stripformer~\cite{Stripformer}, FFTformer~\cite{fftformer} and UHDformer~\cite{uhdformer}. As shown in Tab.~\ref{tab:UHDBlur}, our method achieved a 0.61 dB improvement in PSNR compared to the current state-of-the-art method, UHDformer. Additionally, you can find the visual comparison results in Fig.~\ref{fig:blur}.

\subsubsection{UHD Image Demoiréing Results.}
\begin{figure*}[t!]
    \centering
    \includegraphics[width=\linewidth]{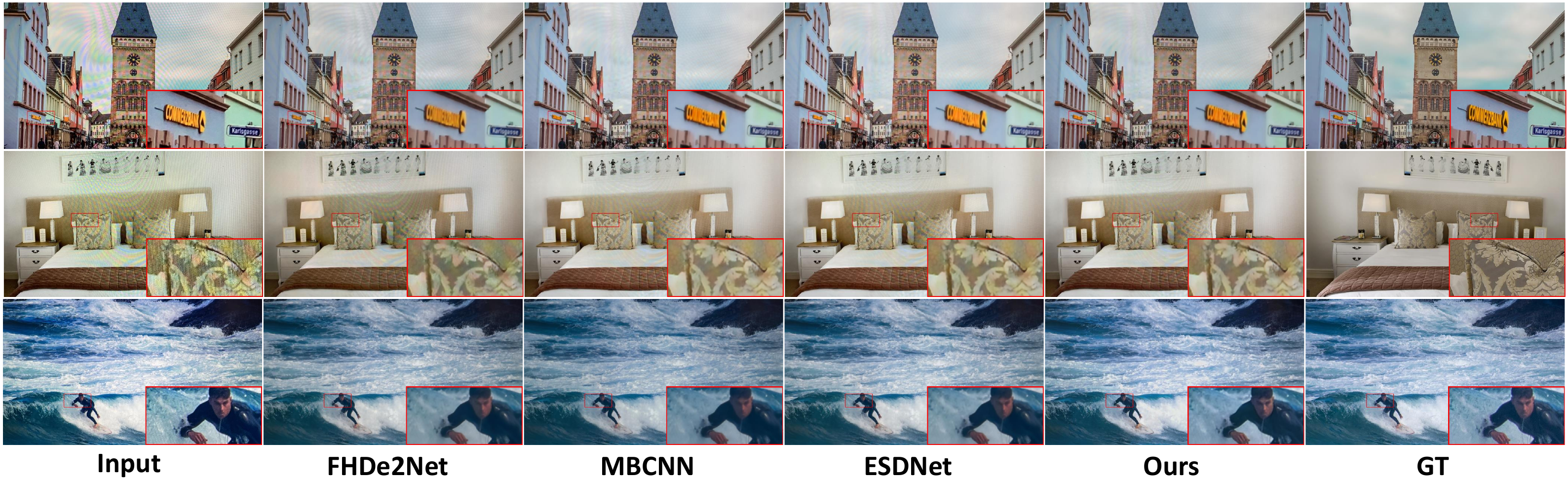}
    \caption{Visual quality comparisons with state-of-the-art methods on UHDM dataset. Please zoom in for details.}
    \label{fig:moire}
\end{figure*}

To validate the effectiveness of our method on the task of moiré pattern removal, we compare it with many methods, including MDDM~\cite{MDDM}, MopNet~\cite{MopNet}, WDNet~\cite{WDNet}, MBCNN~\cite{MBCNN}, FHDe2Net~\cite{FHDe2Net} and ESDNet~\cite{ESDNet}. As shown in Tab.~\ref{tab:UHDM}, our method achieves a 0.23 dB improvement in PSNR compared to the ESDNet, which is also specifically designed for UHD image demoiréing. You can find the comparison of visualization results in Fig.~\ref{fig:moire}.

\begin{table}[t!]
\resizebox{\linewidth}{!}{
\begin{tabular}{c|c|c|cc|c}
\toprule[0.5mm]
Methods   & Type                                          & Venue    & PSNR $\uparrow$  & SSIM $\uparrow$ & Param $\downarrow$ \\ \hline
MDDM  & \multicolumn{1}{c|}{\multirow{5}{*}{non-UHD}} & ICCVW'19 & 20.09 & 0.744 & 7.63M\\
MopNet     & \multicolumn{1}{c|}{}                         & ICCV'19   & 19.48 & 0.757 & 58.5M\\
WDNet      & \multicolumn{1}{c|}{}                         & ECCV'20  & 20.36 & 0.649 &  3.36M\\
MBCNN   & \multicolumn{1}{c|}{}                         & CVPR'20  & 21.41 & 0.793 & 14.19M\\
FHDe2Net & \multicolumn{1}{c|}{}                         & ECCV'20  & 20.33 & 0.749 & 13.5M\\ \hline
ESDNet      & \multicolumn{1}{c|}{\multirow{2}{*}{UHD}}     & ECCV'22    & \underline{22.11} & \underline{0.795} & 5.93M\\
Ours      & \multicolumn{1}{c|}{}                         & -        & \textbf{22.34} & \textbf{0.799} & 7.77M\\
\bottomrule[0.5mm]
\end{tabular}}
\caption{Comparison of quantitative results on UHDM dataset. Best and second best values are indicated with \textbf{bold} text and \underline{underlined} text respectively.}
\label{tab:UHDM}
\end{table}

\subsection{Ablation Study}
We further conduct extensive ablation studies to better understand and evaluate each component in the proposed MixNet. For fair comparisons with the designed baselines, we implement all experiments with the same setting on UHD-LOL4K dataset.

\begin{table}[t!]
\centering
\resizebox{\linewidth}{!}{
\begin{tabular}{c|ccc|cc|c}
\bottomrule[0.5mm]
\#  & GFML       & LFML       & FFL        & PSNR $\uparrow$  & SSIM $\uparrow$  & Param $\downarrow$ \\ \hline
(a) &            &            &            & 30.73 & 0.968 & 7.178M    \\
(b) & \checkmark &            &            & 34.16 & 0.985 & 7.475M    \\
(c) &            & \checkmark &            & 33.93 & 0.984 & 6.882M    \\
(d) &            &            & \checkmark & 32.52 & 0.976 & 7.769M    \\
(e) & \checkmark &            & \checkmark & 36.44 & 0.990 & 8.066M    \\
(f) &            & \checkmark & \checkmark & 34.47 & 0.987 & 7.474M    \\
(g) & \checkmark & \checkmark &            & 35.80 & 0.988 & 7.179M    \\
(h) & \checkmark & \checkmark & \checkmark & \textbf{39.22} & \textbf{0.992} & 7.770M    \\
\bottomrule[0.5mm]
\end{tabular}}
\caption{Ablation study of proposed blocks. In order to ensure consistency in parameter magnitude, we gradually replace the proposed module with ResBlock.}
\label{table:abla}
\end{table}
\begin{table}[t!]
\centering
\resizebox{.6\linewidth}{!}{
\begin{tabular}{r|ccc}
\bottomrule[0.5mm]
     & Stage1 & Stage2 & Stage3 \\ \hline
PSNR $\uparrow$ & 37.94  & 38.57  & 39.22  \\
SSIM $\uparrow$ & 0.989  & 0.990  & 0.992  \\ \bottomrule[0.5mm]
\end{tabular}}
\caption{Ablation study of the number of permute stages. We gradually remove some stages to verify their necessity.}
\label{table:stage}
\end{table}

\subsubsection{Effectiveness of Proposed Modules.}
To demonstrate the gains of proposed modules, we gradually replace the proposed modules with the Residual Block (ResBlock)~\cite{he2016deep} of comparable parameters, and the experimental results in Tab.~\ref{table:abla}. Based on all the results in the Tab.~\ref{table:abla}, all the key designs contribute to the best performance of the full model. Without the GFML (\#a, \#c, \#d, \#f), the model shows the most severe performance degradation. This aligns with our hypothesis, as the excellent long-range dependency modeling capability brought by GMFL is one of the key factors contributing to the outstanding performance of MixNet. And the result suggests the importance of long-range dependency modeling among features. From the results of \#g and \#e, we find that the impact of FFL on performance is higher than that of GFML. We believe that this might be the ability of the FFL to learn compact feature representations, which the ResBlock lacks. And any CNN is capable of learning local features. From the results of \#b, \#c and \#d, we observe that when using the block we proposed alone, GFML and LFML have a higher impact on the results than FFL. This may be because when the feature representation is not effective enough (not from GFML and LFML), learning its compact representation has little effect.

\subsubsection{Effectiveness of the Number of Permute Stages.} 
GFML is the core part of our method. As can be clearly seen in Fig.~\ref{fig:mixer}, its main process can be divided into three steps. Here we verify the necessity of these three stages. In Tab~\ref{table:stage}, we can see that the network's performance drops sharply when some stages are removed. We believe this is because when part of the permute operation is removed, the global modeling capability of GFML is affected, and the remaining permute operation is not enough to complete the long-range dependency modeling of the features.
\begin{table}[t!]
\centering
\resizebox{.9\linewidth}{!}{
\begin{tabular}{r|ccccc}
\toprule[0.5mm]
      & UHD   & LLFormer & UHDFour & UHDformer & Ours  \\ \hline
Param $\downarrow$ & 34.5M & 24.52M   & 17.54M  & 0.34M     & 7.70M \\
RT $\downarrow$   & 1.58s & 7.12s    & 1.67s   & 5.85s     & 0.53s \\ 
\toprule[0.5mm]
\end{tabular}}
\caption{Trainable parameters and running time of some UHD image restoration methods. We record the time cost required to perform inference on $3840 \times 2160$ size images using a single NVIDIA GeForce RTX 3060 GPU.}
\label{tab:time}
\end{table}

\begin{figure}[t!]
    \centering
    \includegraphics[width=.95\linewidth]{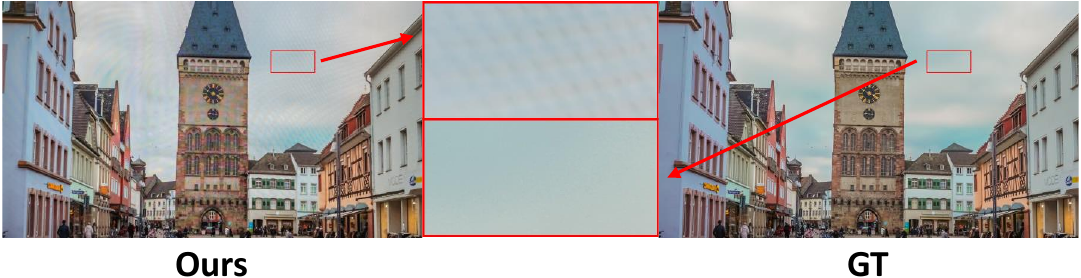}
    \caption{Examples of unsatisfactory restorations. Please zoom in for details.}
    \label{fig:fail}
\end{figure}

\section{Conclusion}
In this paper, we propose a simple yet effective framework called MixNet to solve the efficient UHD image restoration problem. MixNet has a series of FMBs which mainly contain a GFML, LFML, and FFL. GFML explores long-range dependency modeling upon the simple dimensional transformation operation. By permuting the feature maps, GFML associates features from different views. LFML and FFL are designed to capture local features and transform features into a compact representation, respectively. Extensive experiments on both synthetic and real-world datasets showed that the proposed MixNet is more efficient than state-of-the-art methods while achieving competitive performance. 

\subsubsection{Limitations.}
Although our proposed MixNet achieves satisfactory performance across multiple tasks, its efficacy remains limited when dealing with real-world datasets containing complex degradation patterns. For example, in the task of moiré pattern removal, as shown in Fig.~\ref{fig:fail}, the complex moire patterns have not been effectively removed, and the visual results still have a certain gap from the ground truth.
\bibliography{aaai25}

\end{document}